\documentclass[10pt, a4paper]{article}
\usepackage{lrec2022} 
\usepackage{multibib}
\newcites{languageresource}{Language Resources}
\usepackage{graphicx}
\usepackage{tabularx}
\usepackage{soul}
\usepackage{float}
\usepackage{booktabs}
\usepackage{multirow}
\usepackage{titlesec}

\titleformat{\section}{\normalfont\large\bfseries\center}{\thesection.}{1em}{}
\titleformat{\subsection}{\normalfont\SmallTitleFont\bfseries\raggedright}{\thesubsection.}{1em}{}
\titleformat{\subsubsection}{\normalfont\normalsize\bfseries\raggedright}{\thesubsubsection.}{1em}{}
\renewcommand\thesection{\arabic{section}}
\renewcommand\thesubsection{\thesection.\arabic{subsection}}
\renewcommand\thesubsubsection{\thesubsection.\arabic{subsubsection}}

\usepackage{epstopdf}
\usepackage[utf8]{inputenc}

\usepackage{hyperref}
\usepackage{xstring}

\usepackage{color}
\usepackage{xtab}
\usepackage{dblfloatfix}

\DeclareRobustCommand{\indicWords}[1]{%
  \raisebox{-\dp\strutbox}{%
    \includegraphics[page=\csname indicWords#1\endcsname]{indicWords}%
  }%
}

\title{HiNER: A Large Hindi Named Entity Recognition Dataset}

\name{Rudra Murthy\textsuperscript{2}, 
{\bf \large Pallab Bhattacharjee\textsuperscript{1}, } 
{\bf \large Rahul Sharnagat\textsuperscript{3},} \\
{\bf \large Jyotsana Khatri\textsuperscript{1},} 
{\bf \large Diptesh Kanojia\textsuperscript{4},}
{\bf \large Pushpak Bhattacharyya\textsuperscript{1}}}

\address{
\textsuperscript{1}CFILT Lab, IIT Bombay, India. \\
\textsuperscript{2}IBM IRL, Bangalore, India.
\textsuperscript{3}Walmart Labs, USA. \\
\textsuperscript{4}Surrey Institute for People-centred AI, University of Surrey, United Kingdom. \\
\textsuperscript{1}bhattacharjee.pallab9@gmail.com, \{jyotsanak, pb\}@iitb.ac.in\\
\textsuperscript{2}rmurthyv@in.ibm.com, \textsuperscript{3}rdsharnagat@gmail.com, \textsuperscript{4}d.kanojia@surrey.ac.uk
}

\begin{document}
\abstract{
Named Entity Recognition (NER) is a foundational NLP task that aims to provide class labels like \textit{Person}, \textit{Location}, \textit{Organisation}, \textit{Time}, and \textit{Number} to words in free text. Named Entities can also be multi-word expressions where the additional I-O-B annotation information helps label them during the NER annotation process. While English and European languages have considerable annotated data for the NER task, Indian languages lack on that front- both in terms of quantity and following annotation standards. This paper releases a significantly sized standard-abiding Hindi NER dataset containing 109,146 sentences and 2,220,856 tokens, annotated with 11 tags. We discuss the dataset statistics in all their essential detail and provide an in-depth analysis of the NER tag-set used with our data. The statistics of tag-set in our dataset show a healthy per-tag distribution, especially for prominent classes like \textit{Person}, \textit{Location} and \textit{Organisation}. Since the proof of resource-effectiveness is in building models with the resource and testing the model on benchmark data and against the leader-board entries in shared tasks, we do the same with the aforesaid data. We use different language models to perform the sequence labelling task for NER and show the efficacy of our data by performing a comparative evaluation with models trained on another dataset available for the Hindi NER task. Our dataset helps achieve a weighted F1 score of 88.78 with all the tags and 92.22 when we collapse the tag-set, as discussed in the paper. To the best of our knowledge, no available dataset meets the standards of volume (amount) and variability (diversity), as far as Hindi NER is concerned. We fill this gap through this work, which we hope will significantly help NLP for Hindi. We release this \href{https://github.com/cfiltnlp/HiNER}{dataset with our code and models} for further research.\\ \newline }

\maketitleabstract
\section{Introduction}

Named Entity Recognition (NER) is an essential lower-level task~\cite{DBLP:journals/corr/MaH16} in Natural Language Processing (NLP), used to extract and categorize naming entities into a predefined set of classes such as \textit{person}, \textit{location}, \textit{organization}, \textit{numeral} and \textit{temporal entities}. A well-performing NER system can help the downstream tasks of Machine Translation~\cite{babych2003improving}, Information Extraction~\cite{neudecker2016open}, and Questions Answering~\cite{moldovan2002role}. With the recent surge in the NER research~\cite{sohrab-miwa-2018-deep,plank-2019-neural,copara-etal-2020-contextualized,grancharova-dalianis-2021-applying}, the NLP community has also created large annotated datasets for the NER task~\cite{ali-etal-2020-siner,ding2021few} including code-mixed datasets~\cite{singh-etal-2018-named}. Research in NER has seen remarkable progress since the early approaches and evaluation metrics proposed by~\newcite{sang2002memory,sang2003introduction}. The task of NER belongs to the class of NLP problems, which can be modelled as a `sequence labelling' problem akin to the tasks of Part-of-Speech (PoS) tagging and chunking. With the advent of deep learning-based approaches, sequence labelling tasks have invited much attention with successful methods like BiLSTM-CRF~\cite{DBLP:journals/corr/HuangXY15} and Transformers architecture-based fine-tuning~\cite{vaswani2017attention,wolf2019huggingface}. However, these methods require significant data to produce a well-performing NER system for any language. 

\begin{table}[!h]
\resizebox{\columnwidth}{!}{%
\begin{tabular}{@{}lrrrr@{}}
\toprule
 \multicolumn{1}{l}{} & \textbf{Ours} & \textbf{Wiki ANN} & \textbf{Fire 2014} & \textbf{IJCNLP 2008} \\ \midrule
 Sentences & 109146 & 7000 & 9622 & 21833 \\
 Tokens & 2220856 & 41256 & 116103 & 541682  \\
 \midrule
 Person & 37605 & 22959 & 2112 & 4235 \\
 Location & 198282 & 20131 & 2268 & 4307 \\
 Organization & 26509 & 14204 & 170 & 1272 \\ 
 \bottomrule
\end{tabular}%
}
\caption{Comparison of HiNER data statistics with existing Hindi NER datasets}
\label{tab:datasetStatsCompare}
\end{table}

NLP for Indian languages has shown progress with the availability of large language models~\cite{kumar-etal-2020-passage,kakwani2020inlpsuite,khanuja2021muril} which can help perform various NLP tasks. However, there has been little progress in terms of producing NER datasets for Indian languages, especially for Hindi, which approximately 342 million people speak across the world\footnote{\href{https://en.wikipedia.org/wiki/List_of_languages_by_total_number_of_speakers}{Wikipedia: List of Language by Speakers}}. NER systems trained on our dataset are expected to perform better than the existing systems trained on lesser data. Existing datasets are either much smaller or have been automatically annotated (silver standard), rendering them incapable of performing the NER task with high accuracy. Moreover, during the creation of a Hindi NER system, one faces various linguistic challenges like:
\begin{description}
\item[No Capitalization:] Unlike English or other languages which use the Latin script, Hindi does not have capitalization as a feature which should have been helpful for performing the NER task,
\item[Ambiguity:] Proper nouns in Hindi can be ambiguous as the same word can belong to a different PoS category. For example, a common Indian female name like `\textit{Pushpa}' can be both a proper noun and a common noun meaning `flower',
\item[Spelling Variations:] The spelling of some words in Hindi can differ depending on the local region in India. For example, the concept or sense of `Plant' can be denoted by both the words- `\textit{vanaspati}' and `\textit{banaspati}', 
\item[Free Word Order:] Languages like Hindi, which follow a free word order, make the NER task more challenging as computational approaches can not be complemented with a pattern of PoS tags, or strict word order.
\end{description}

Due to the challenges discussed above, it is imperative to train Hindi NER models with a sizeable human-annotated dataset so that deep learning-based approaches can generalize and perform well. 

This paper describes our longstanding efforts toward creating a sizeable human-annotated dataset for Hindi NER, which we call ``HiNER''. We collect this dataset with the help of one annotator and perform experiments to evaluate the efficacy of various deep learning-based approaches. We also include the current public datasets in these experiments and compare the performance of these approaches across datasets. Our work also describes the NER tool developed in-house to help our annotators. This tool also provides a NER service on the back-end, which helps tag the NER data initially, and allows our annotators to post-edit the NER tags with ease. We describe the creation of the back-end NER engine in detail. We also discuss our dataset regarding the various sources and domains and provide an in-depth analysis of the NER tag-set we use for our dataset. The contributions of this work are summarized below:
\begin{itemize}
    \item We collect a large manually annotated NER dataset for Hindi (HiNER) and release it publicly.
    \item We evaluate the performance of various deep learning-based NER approaches on our dataset and compare the performance with other publicly available datasets.
    \item We also release our data, code and models.
\end{itemize}

\section{Related Work}

For the task of Named Entity Recognition, much pre-existing literature attempts to solve the problem in different languages and domains. However, in this section, we discuss existing literature for Hindi and other Indian languages. We also describe research that highlights different approaches for the NER task. The IJCNLP 2008 NER dataset comprises NER data in five languages, namely Hindi, Bengali, Oriya, Telugu, and Urdu~\cite{ijcnlp-2008-ijcnlp-08}. This data has been used extensively in previous research for the Hindi NER task~\cite{ekbal2008language,gupta2010think,bhagavatula2012language,gali2008aggregating,saha-etal-2008-hybrid,saha-etal-2008-word}. The FIRE 2014 dataset \cite{FIRE2014} consists of NER data in four languages, namely Hindi, Tamil, Malayalam, and English~\cite{choudhury2014overview}. Similarly, the WikiANN data~\cite{pan2017cross} consists of NER data in 282 languages, including Hindi; however, it is tagged automatically and a known `silver-standard' dataset for the NER task. Moreover, it consists of only 10000 sentences in total.~\newcite{rahimi-etal-2019-massively} utilise transfer learning for multilingual NER and discuss their results for 41 languages in zero-shot, few-shot and high-resource scenarios. ~\newcite{singh-etal-2018-named} use Long Short Term Memory (LSTM), Decision Trees, and Conditional Random Fields (CRF) to perform the NER task on code-mixed Hindi-English social media text. Past research has also tried to utilise voting algorithm-based hybrid approaches, which take CRF, Maximum Entropy (MaxEnt) and rules into  account~\cite{srivastava2011named}. The authors use the IJCNLP-08 dataset for Hindi, and their approach achieved 82.95 as the F-score.~\newcite{gupta2010think} also identify a local context within the global information for the task of Hindi NER and report a performance gain of about 10\% resulting in a 72\% F1 score. 

Recent work on Indian language NER utilises various deep learning-based approaches for the task.~\newcite{9510090} utilise a Bidirectional LSTM (BiLSTM) architecture with the help of contextualized ELMo word representations~\cite{peters-etal-2018-deep}. Similarly, for the Hindi NER task,~\newcite{athavale-etal-2016-towards} explore the use of BiLSTM and utilise multiple datasets to report around 77.48\% F1 score for all tags. Among multilingual approaches, past research has attempted to utilise morphological and phonological sub-word representations to help the NER task for four languages, including Hindi~\cite{chaudhary2018adapting}. ~\newcite{c-s-lalitha-devi-2020-deeper} also proposes various typological features and proposes a machine learning-based approach for the NER task in many language families.~\cite{10.1145/3238797,murthy-etal-2018-judicious} demonstrate on FIRE 2014 data that training with combined labelled data of multiple languages can help in Indian language NER. With the help of non-speaker annotations, ~\newcite{tsygankova2020building} show that even without the help of native speakers of the language, manual annotation for a NER task helps perform better than the available cross-lingual methods, which use modern contextualised representations. Focusing on the challenge of code-switching in NER data,~\newcite{aguilar-etal-2020-lince} propose a new benchmark for code-switching they call LinCE and perform experiments for Hindi-English code-switched data to show encouraging results (75.96\% F1). As discussed earlier, there is past research on NER, including NER for the Hindi language, but there are not sufficiently large datasets for the task of Hindi NER. With this paper, we release a large NER dataset collected \textbf{over several years with the help of a single annotator} and show its efficacy with the use of various available language models for Indian languages.

\section{Dataset Creation}

In this section, we discuss the creation of our dataset in detail. We follow the same guidelines as the CoNLL-2003 NER Shared task \cite{TjongKimSang:2003:ICS:1119176.1119195}. The CoNLL-2003 NER Shared task\footnote{\url{https://www.clips.uantwerpen.be/conll2003/ner/annotation.txt}} contains the following tags: \textit{Person}, \textit{Location}, \textit{Organization}, and \textit{MISC}. While the CoNLL-2003 data does not contain \textit{TIMEX} and \textit{NUMEX} tags, these tags are part of Onto-notes \cite{pradhan-etal-2013-towards} and we add \textit{TIMEX} and \textit{NUMEX} tags as part of our tag-set. During the annotation process, we observed the \textit{MISC} tag to be too coarse and further include the \textit{Language}, \textit{Game}, \textit{Literature}, \textit{Religion}, and \textit{Festival} as separate tag entities. We believe these fine-grained tags help create a more detailed NER dataset, thus helping the computational models be more accurate with the NER task. Some of challenges in Hindi NER is there is no capitalization, no use of camel case for names, and the free word order.
In the following sub-section, we discuss the statistics of this dataset in detail. As one annotator has annotated the data, we can not provide any inter-annotator agreement details with our paper. 

\begin{table}[!t]
\centering
\resizebox{\columnwidth}{!}{%
\begin{tabular}{@{}lrr|c@{}}
\toprule
\textbf{Data Splits} & \#sentences & \#words & Split Size \\ \midrule
Training & 76025 & 1382979 & 70\% \\
Development & 10861 & 200259 & 10\% \\
Testing & 21722 & 553961 & 20\% \\
\midrule
\textbf{Total} & 108,608 & 2,137,199 & 100\% \\
\bottomrule
\end{tabular}%
}
\caption{\textbf{HiNER} dataset statistics in terms of the number of sentences (\#sentences), number of words (\#words), and the splits created for the Hindi NER task}
\label{tab:dataSplit}
\end{table}

\subsection{Dataset Statistics}

We annotate data from the ILCI Tourism domain~\cite{jha} and a subset of the `news' domain corpus from \newcite{goldhahn-etal-2012-building}. We take a subset of $9989$ sentences out of $25,000$ sentences from the ILCI tourism domain and the rest from the news domain. The dataset includes a total of 108,608 sentences. The number of entities for each tag is shown in Table~\ref{tab:dataStats} for a total of 11 tags. We also show the statistics of FIRE 2014 dataset in Table~\ref{tab:datasetStatsCompare} for comparison.

\begin{table}[!ht]
\centering
\begin{tabular}{@{}lrrr|r@{}}
\toprule
 & \multicolumn{1}{c}{\textbf{Train}} & \multicolumn{1}{c}{\textbf{Dev}} & \multicolumn{1}{c}{\textbf{Test}} & \multicolumn{1}{|c}{\textbf{Total}}  \\ \midrule
PERSON & 26310 & 3771 & 7524 & 37605\\ \midrule
LOCATION & 137995 & 20100 & 40187 & 198282\\ \midrule
NUMEX & 17194 & 2555 & 4662 & 24411\\ \midrule
ORGANIZATION & 18508 & 2645 & 5356 & 26509\\ \midrule
MISC & 4070 & 553 & 1080 & 5703\\ \midrule
LANGUAGE & 4187 & 571 & 1190 & 5948\\ \midrule
GAME & 1214 & 180 & 369 & 1763\\ \midrule
TIMEX & 13047 & 1762 & 3653 & 18462\\ \midrule
RELIGION & 823 & 133 & 234 & 1190\\ \midrule
LITERATURE & 597 & 74 & 181 & 852\\ \midrule
FESTIVAL & 203 & 30 & 40 & 273\\ \midrule
\textbf{Total} & 224148 & 32374 & 64476 & 320998\\
\bottomrule
\end{tabular}%
\caption{Number of Entity mentions (Phrases) in \textit{Train}, \textit{Dev}, \textit{Test} splits for the \textbf{HiNER dataset}}
\label{tab:dataStats}
\end{table}

\begin{figure}[!ht]
    \centering
    \includegraphics[width=1\columnwidth]{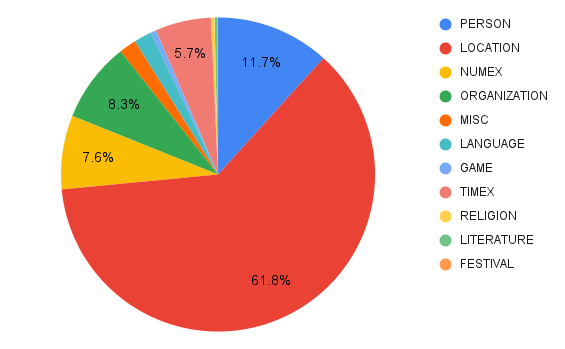}
    \caption{HiNER Tagset Details}
    \label{fig:dataset}
\end{figure}

\subsection{NER Tool}

\begin{figure*}[t!]
    \centering
    \includegraphics[width=0.9\textwidth]{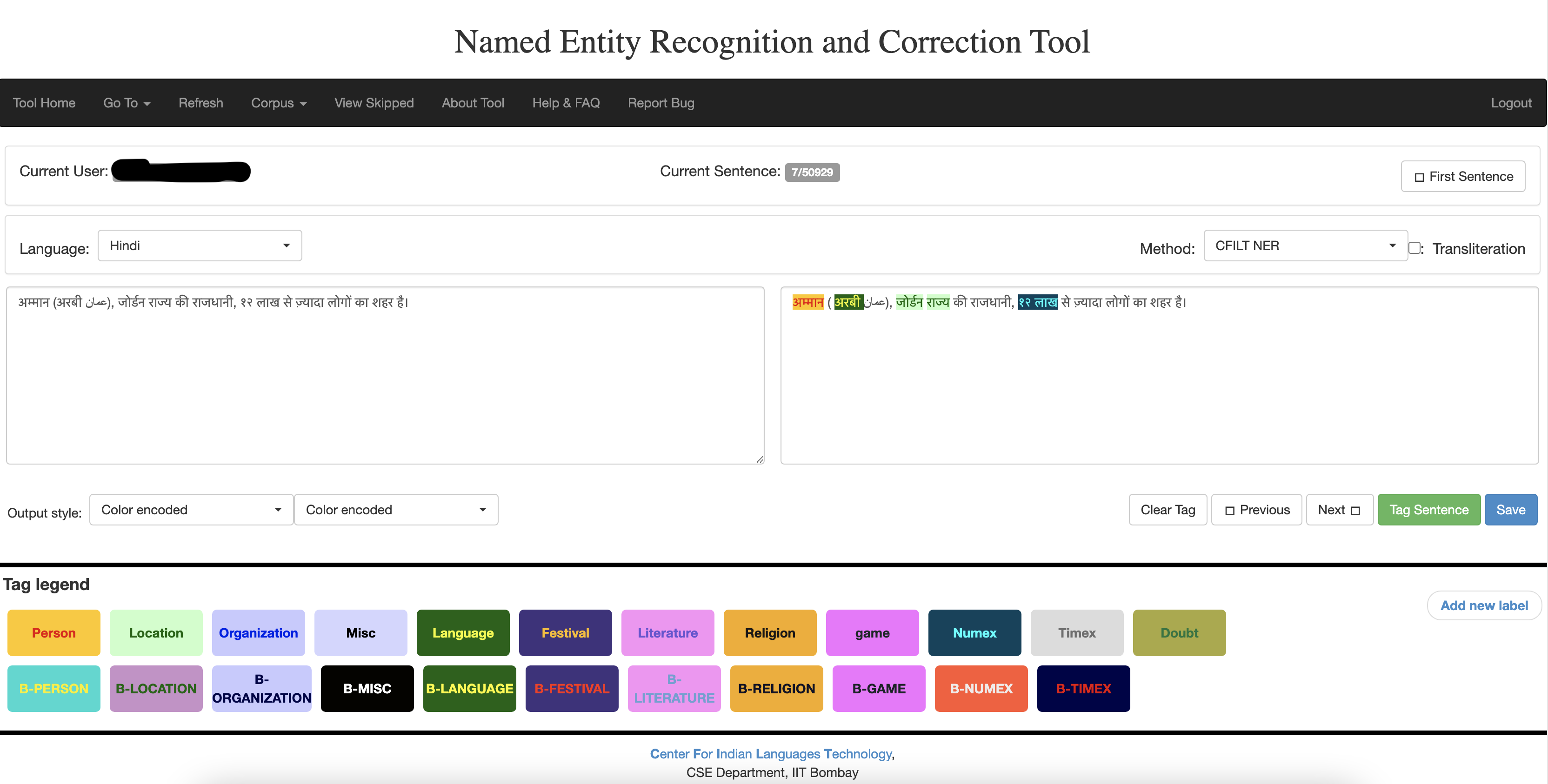}
    \caption{NER Tool Screenshot showing the tool interface with redacted user name to preserve anonymity. }
    \label{fig:screenshot}
\end{figure*}

\begin{table*}[!htb]
\centering
\begin{tabular}{@{}p{8.2cm}p{7cm}@{}}
\toprule
\multicolumn{1}{c}{\textbf{Sentence}} & \multicolumn{1}{c}{\textbf{Confusion and Resolution}} \\ \midrule
\indicWords{Asentence} & Will \textit{high court} be tagged as Organization or  \\
\textbf{uchcha nyAyAlaya} rAjya kI sarvopari nyAyika sattA hai , & not? Every state in India has a high court.  \\
\textbf{high court} state's supreme judicial authority is & Since high court doesn't refer to a particular high \\
\textbf{High Court} is the supreme judicial authority of the state. & court as in \textit{Bombay High Court}, we leave it untagged. \\
\midrule
\indicWords{Bsentence} & Should \textit{chitra ChAyA} be tagged as \textit{Organization} \\
\textbf{chitra ChAyA} hindI kI eka philmI mAsika patrikA hai &  or \textit{Literature}? In the context provided, \textit{chitra} \\
chitra ChAyA hindi's one filmy monthly periodical is &  \textit{ChAyA} refers to the periodical and not the \\
\textbf{Chitra Chhaya} is a Hindi film monthly magazine. & organization, we tag them as \textit{Literature} \\

\midrule
\indicWords{Csentence} & Will \textit{garuDa} be tagged as \textit{Person} name or not   \\
viShNu ke vAhana \textit{garuDa} haiM & a named entity? Here, \textit{garuDa} is the name of \\
viShNu's vehicle \textit{garuDa} is & an eagle from Hindu mythology. According \\
\textbf{Garuda} is the vehicle of Vishnu. & to CoNLL 2003 entity guidelines, it should be tagged as \textit{Person} entity \\
\midrule
\indicWords{Dsentence} & What should be the entity label of \textit{chaTanI} \\ 
rikkI jaya (samrAja jayamaMgala) TriniDADa ke \textbf{chaTanI saMgIta} kA kalAkAra hai  & \textit{saMgIta} ? Here \textit{chaTanI saMgIta} refers to a fusion genre of Indian folk music, specifically Bhojpuri  \\
rikkI jaya (samrAja jayamaMgala) Trinidad's \textit{chaTanI saMgIta} artist is &  folk music, with local Caribbean calypso and soca music, and later on Bollywood music. Hence, will \\
Rikki Jai (Emperor Jayamangala) is a Trinidadian \textbf{chutney music} artist. & be tagged as \textit{Misc} entity.\\
\bottomrule
\end{tabular}
\caption{Sentences flagged by the annotator with the entity highlighted and the reasoning for the final decision.}
\label{tab:ambiguousSentence}
\end{table*}

To ease the annotation task, we create an online tool based on PaCMan~\cite{kanojia-etal-2014-pacman}. We modify the architecture of PaCMan to allow the upload of untagged NER data~\footnote{\href{http://www.cfilt.iitb.ac.in/ner/admin/login.php
}{Link: Tool Interface}}. Further, we make changes in the tool front-end to show the full tag-set on the source and target sides of the screen as shown in Figure~\ref{fig:screenshot}. The untagged data is also shown on the left side of the screen in a text box for clarity to the annotator; however, the annotator must tag the sentence on the right side. Borrowing a feature from the PaCMan interface, we modify the customized right-click-based context menu for different NER tags. The annotator must go through the sentence manually, highlight the named entity and then right-click to provide it with the correct label. This simplified annotation process allows our annotators to label the data with ease. The tool stores the data on a MySQL-based back-end and allows for downloading data files from the interface. Each time an annotator progresses onto the following sentence, the previously tagged sentence is saved automatically. The tool also saves the annotation state in the database, thus allowing an annotator to arrive at the next untagged instance in the database when they log on later. We further simplify the annotation by providing them with a baseline NER engine that allows them to tag the sentence initially and simply ``post-edit'' the annotations and save the correctly labelled sentence. We describe this baseline NER engine in the following subsection. 

\begin{table*}[!b]
\centering
\resizebox{0.97\textwidth}{!}{%
\begin{tabular}{@{}lrrrrr@{}}
\toprule
\multicolumn{1}{c}{\textbf{}} & \multicolumn{1}{c}{\textbf{Indic-BERT}} & \multicolumn{1}{c}{\textbf{mBERT}} & \multicolumn{1}{c}{\textbf{MuRIL}} & \multicolumn{1}{c}{\textbf{XLM-R\textsubscript{base}}} & \multicolumn{1}{c}{\textbf{XLM-R\textsubscript{large}}} \\
\midrule
Festival & 9.52 $\pm$ 11.90 & 8.57 $\pm$ 17.14 & 0.00 $\pm$ 0.00 & 11.34 $\pm$ 14.53 & \textbf{46.73 $\pm$ 23.96} \\
Game & 50.05 $\pm$ 8.33 & 50.92 $\pm$ 20.52 & 40.88 $\pm$ 22.96 & 47.57 $\pm$ 10.63 & \textbf{59.63 $\pm$ 7.94} \\
Language & 89.22 $\pm$ 1.15 & 90.07 $\pm$ 1.13 & 90.08 $\pm$ 1.02 & 90.64 $\pm$ 0.56 & \textbf{91.42 $\pm$ 0.57} \\
Literature & 21.64 $\pm$ 26.12 & 53.56 $\pm$ 10.93 & 44.23 $\pm$ 22.17 & 40.54 $\pm$ 23.39 & \textbf{56.69 $\pm$ 6.32} \\
Location & 94.10 $\pm$ 0.56 & 93.92 $\pm$ 0.57 & 94.81 $\pm$ 0.37 & 94.07 $\pm$ 0.76 & \textbf{94.86 $\pm$ 0.40} \\
Misc & 56.14 $\pm$ 10.97 & 61.24 $\pm$ 10.99 & 62.84 $\pm$ 4.22 & 60.38 $\pm$ 12.19 & \textbf{67.86 $\pm$ 2.19} \\
NUMEX & 65.56 $\pm$ 3.25 & 67.21 $\pm$ 1.50 & 68.31 $\pm$ 1.77 & 66.72 $\pm$ 2.32 & \textbf{69.10 $\pm$ 0.95} \\
Organization & 76.68 $\pm$ 1.33 & 74.81 $\pm$ 3.10 & 78.26 $\pm$ 2.46 & 76.02 $\pm$ 2.73 & \textbf{78.76 $\pm$ 1.70} \\
Person & 83.65 $\pm$ 0.50 & 81.10 $\pm$ 1.70 & 84.60 $\pm$ 1.30 & 83.04 $\pm$ 0.86 & \textbf{85.14 $\pm$ 0.94} \\
Religion & 65.94 $\pm$ 3.20 & 68.55 $\pm$ 7.58 & 53.43 $\pm$ 26.74 & 67.70 $\pm$ 5.78 & \textbf{72.27 $\pm$ 2.68} \\
TIMEX & 80.20 $\pm$ 1.11 & 81.15 $\pm$ 1.24 & 81.17 $\pm$ 1.20 & 79.50 $\pm$ 0.85 & \textbf{80.63 $\pm$ 1.05} \\
\midrule
Micro & 87.44 $\pm$ 0.62 & 87.11 $\pm$ 1.01 & 88.27 $\pm$ 0.92 & 87.36 $\pm$ 1.09 & \textbf{88.73 $\pm$ 0.60} \\
Macro & 62.97 $\pm$ 5.19 & 66.46 $\pm$ 5.93 & 63.51 $\pm$ 7.36 & 65.23 $\pm$ 6.04 & \textbf{73.01 $\pm$ 3.35} \\
Weighted & 87.25 $\pm$ 0.88 & 87.06 $\pm$ 1.28 & 88.27 $\pm$ 1.08 & 87.29 $\pm$ 1.23 & \textbf{88.78 $\pm$ 0.57} \\
\bottomrule\\
\end{tabular}%
}
\caption{Test Set F1-Score of various pre-trained LMs on our HiNER dataset. This table reports a mean F1-score and its standard deviation over 5 runs.}
\label{tab:resultsOurDatasetLabel}
\end{table*}

\subsection{NER Engine}

We developed a NER engine to provide Named Entity suggestions to our annotators. Each sentence from our dataset is presented on the tool interface as shown in the screenshot (Figure~\ref{fig:screenshot}), and a button (``Tag Sentence'') which allows the NER engine to perform NE tagging of the sentence on the back-end. The tagged sentence is shown to the annotator on the annotation screen's right side, which can be edited later. Our annotators reported that they could easily modify the tool's engine errors. This NER engine was developed using FIRE 2013 Hindi NER corpus~\cite{FIRE2013}. Due to the limited size of the training corpus, it was hard to create a tagger that could learn a generic sequence of tags. To support the model, we employed word2vec~\cite{word2vec} to learn the semantic embeddings for single and multi-word tokens based on a large Hindi Wikipedia dump. These learned embeddings were then used to train a simple perceptron-based neural network model to infer named entities. A separate service was created in conjunction with the front-end UI of our NER tool to handle the annotation requests. Our annotators reported that this engine was prone to errors, especially when tagging multi-word named entities, but it could handle commonly used named entities. 

\subsection{Annotation Ambiguity}
As only one annotator annotated the data, ensuring that the dataset's quality is not compromised is essential. We encouraged the annotator to raise reports for entities he was not confident in tagging. The authors then take a majority voting on such instances to assign an appropriate entity or not an entity label. We now provide a few examples of such instances raised by the annotator in Table~\ref{tab:ambiguousSentence}.



\section{Dataset Evaluation}

In this section, we discuss the evaluation of our dataset based on different approaches to NER. With the help of our annotator, we collected the NER-labelled dataset as described above. We perform the task of Hindi NER with the help of various contextual language models and in different settings. With our dataset, we create a data split of 70\% for training, 10\% for development, and 20\% for testing, with statistics, as shown in Table~\ref{tab:dataSplit}. We ensured a balanced percentage of tags in each of the splits with stratification, as can be seen from Table~\ref{tab:dataStats}. 

\begin{table*}[!h]
\centering
\resizebox{0.97\textwidth}{!}{%
\begin{tabular}{lrrrrr}
\toprule
{} & \multicolumn{1}{c}{\textbf{Indic-BERT}} & \multicolumn{1}{c}{\textbf{mBERT}} & \multicolumn{1}{c}{\textbf{MuRIL}} & \multicolumn{1}{c}{\textbf{XLM-R\textsubscript{base}}} & \multicolumn{1}{c}{\textbf{XLM-R\textsubscript{large}}} \\
\midrule
Location & 94.33 $\pm$ 0.63 & 94.44 $\pm$ 0.24 & 94.95 $\pm$ 0.25 & 95.07 $\pm$ 0.20 & 95.06 $\pm$ 0.33 \\
Organization & 78.29 $\pm$ 1.57 & 78.42 $\pm$ 1.13 & 79.87 $\pm$ 0.81 & 79.57 $\pm$ 0.62 & 80.53 $\pm$ 0.40 \\
Person & 84.70 $\pm$ 0.61 & 82.20 $\pm$ 0.98 & 85.66 $\pm$ 0.48 & 85.18 $\pm$ 0.66 & 85.34 $\pm$ 0.54 \\
\midrule
micro avg & 91.37 $\pm$ 0.67 & 91.10 $\pm$ 0.34 & 92.09 $\pm$ 0.27 & 92.06 $\pm$ 0.27 & 92.20 $\pm$ 0.22 \\
macro avg & 85.77 $\pm$ 0.91 & 85.02 $\pm$ 0.66 & 86.83 $\pm$ 0.41 & 86.61 $\pm$ 0.40 & 86.98 $\pm$ 0.22 \\
weighted avg & 91.34 $\pm$ 0.71 & 91.08 $\pm$ 0.35 & 92.11 $\pm$ 0.29 & 92.11 $\pm$ 0.27 & \textbf{92.22 $\pm$ 0.22} \\
\bottomrule
\end{tabular}%
}
\caption{Test Set F1-Score of various pre-trained LMs on our HiNER dataset (Collapsed). This table reports a mean F1-score and its standard deviation over 5 runs.}
\label{tab:collapsedResultsOurDatasetLabel}
\end{table*}

\subsection{Experimental Setup}

With the advent of contextualized word representations, various language models have been proposed which can be utilized to perform NLP tasks~\cite{devlin2018bert,conneau-etal-2020-unsupervised,kakwani2020inlpsuite,khanuja2021muril}. We use these four models to evaluate the performance of the NER task on our dataset. Additionally, we use the FIRE 2014 dataset to compare the efficacy of both datasets and present the results in the next section. We also utilize the models trained on our data and test on the FIRE 2014 test split to evaluate the model performance in a cross-dataset scenario. 

\begin{figure}[!htb]
    \includegraphics[width=0.5\textwidth]{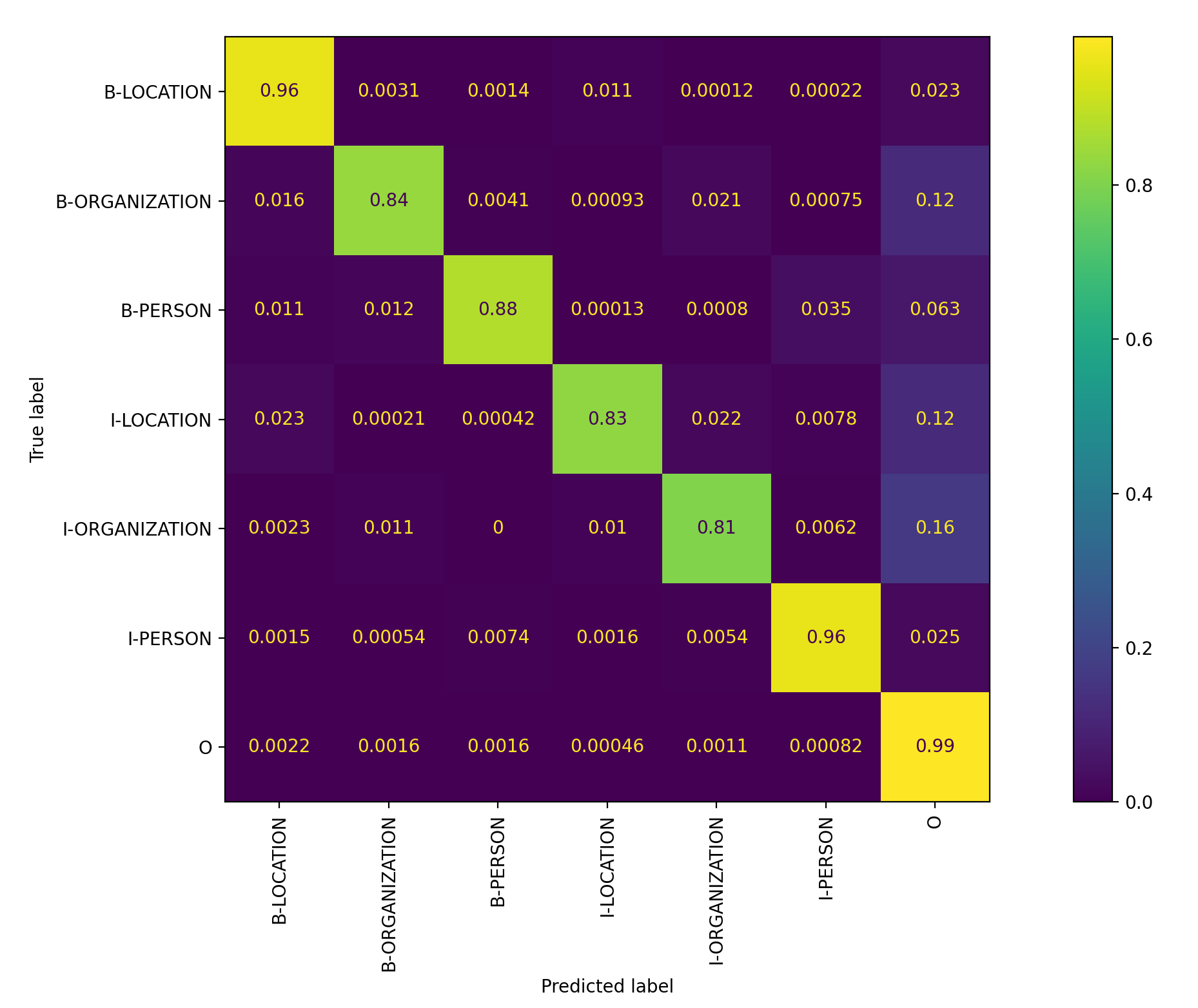}
    \caption{Confusion Matrix on HiNER Collapsed data from XLM-R\textsubscript{large} Model}
    \label{fig:confusionHiNERCollapsed}
\end{figure}

We use the variation mBERT$_{base-cased}$ of multilingual BERT (mBERT), which supports 104 languages, and has 12 layers with 768 hidden layers, along with a total of 110M parameters. We use \textit{XLM-R\textsubscript{base}} and \textit{XLM-R\textsubscript{large}} \cite{conneau-etal-2020-unsupervised} which are pre-trained multilingual language models to fine-tune for NER task. However, IndicBERT \cite{kakwani2020inlpsuite}, and MuRIL \cite{khanuja2021muril} are more suited to the task as it supports Indian languages in particular and is trained on shared vocabulary from Indic languages. IndicBERT is trained on 12 major Indian languages, including Hindi, is trained on around 9 billion tokens, and has a restriction on the maximum sequence length (128). Similarly, MuRIL is a model pre-trained on 17 Indian languages and their transliterated counterparts. 

We perform hyper-parameter tuning of each model and select the hyper-parameters giving the best F-Score on the development set. The model is trained using the best hyper-parameter for $5$ runs. We report the mean and standard deviation over all runs in all our experiments. Considering the batch size and learning rate as hyper-parameters, we provide the following variations for batch size $\{8, 16, 32\}$. Due to GPU memory limitations, we did not experiment with larger batch sizes. Similarly, we vary the learning rate in the following range \{1e-3, 3e-3, 5e-3, 1e-4, 3e-4, 5e-4, 1e-5, 3e-5, 5e-5, 1e-6, 3e-6, 5e-6 \} and select the best learning rate.

We perform experiments on two variations of our dataset: one with all $11$ tags, and the collapsed version with only $3$ tags (\textit{Person}, \textit{Location}, \textit{Organization} tags). For FIRE 2014 data \cite{FIRE2014}, similar to \newcite{10.1145/3238797,murthy-etal-2018-judicious}, we consider only \textit{Person}, \textit{Location}, \textit{Organization} tags. We use the I-O-B encoding as input format for model training and report the results using Seqeval~\cite{seqeval} generate evaluation statistics.



\subsection{Evaluation Results}

From Table~\ref{tab:resultsOurDatasetLabel}, we observe that \textit{XLM-R\textsubscript{large}} performs the best on our dataset followed by \textit{MuRIL}. \textit{Festival} entity has the lowest performance across all models and also has highest standard deviation across different runs compared to other tags. All the models are able to identify \textit{Language}, \textit{Location}, \textit{Person}, and, \textit{TIMEX} entities with relatively high accuracy.

Table \ref{tab:collapsedResultsOurDatasetLabel} reports results on our collapsed data. We focus on \textit{Person}, \textit{Location}, and, \textit{Organization} tags here. Similar to our earlier results, \textit{XLM-R\textsubscript{large}} performs the best followed by \textit{MuRIL} model. Unsurprisingly, \textit{Organization} entity has relatively lowest F1 score compared to \textit{Person} and \textit{Location} entities.

\subsubsection*{Zero-Shot Performance Test}

Table \ref{tab:resultsOurDatasetTransfer} reports the results from our experiments in the Zero-Shot experiment setup. For this set of experiments, we use FIRE 2014 Hindi NER data \cite{FIRE2014}. When we perform a zero-shot transfer from our dataset to the FIRE 2014 dataset, the results are poor compared to training on the FIRE 2014 dataset. Surprisingly, \textit{MuRIL} model performs the best in the zero-shot transfer set-up compared to other pre-trained language models. 

\begin{table}[!t]
\centering
\resizebox{\linewidth}{!}{%
\begin{tabular}{@{}lrrc@{}}
\toprule
\multicolumn{1}{l}{\textbf{Model}} & 
\multicolumn{1}{c}{\begin{tabular}[c]{@{}c@{}}\textbf{HiNER} \end{tabular}} & 
\multicolumn{1}{c}{\textbf{FIRE}} & 
\multicolumn{1}{c}{\begin{tabular}[c]{@{}c@{}}\textbf{HiNER} $\rightarrow$ \textbf{FIRE} \end{tabular}} 
\\ 
 & \multicolumn{1}{c}{\textbf{(collapsed)}} & \multicolumn{1}{c}{\textbf{2014}} & \multicolumn{1}{c}{\textbf{Zero-Shot}} \\
\midrule
IndicBERT & 91.37 $\pm$ 0.67 & 62.79 $\pm$ 0.68 & 46.82 $\pm$ 1.87  \\
mBERT & 91.10 $\pm$ 0.34 & 62.14 $\pm$ 0.59 & 47.60 $\pm$ 1.62 \\
MuRIL & 92.09 $\pm$ 0.27 & 62.58 $\pm$ 2.44 & 55.26 $\pm$ 1.59  \\
XLM-R\textsubscript{base} & 92.06 $\pm$ 0.27 & 65.63 $\pm$ 0.76 & 49.48 $\pm$ 0.79 \\
XLM-R\textsubscript{large} & \textbf{92.20 $\pm$ 0.22} &  \textbf{66.75 $\pm$ 0.30} & \textbf{49.52 $\pm$ 3.12} \\
\bottomrule
\end{tabular}%
}
\caption{Test Set Micro F1-Score of various pre-trained LMs on both datasets where HiNER (collapsed) is our dataset with only the \textit{Person}, \textit{Location} and \textit{Organization} tags. This table reports a mean F1-score and its standard deviation over 5 runs.}
\label{tab:resultsOurDatasetTransfer}
\end{table}

\section{Discussion}
We plot the confusion matrix on HiNER Collapsed data from one of the runs of XLM-R\textsubscript{large} Model in Figure \ref{fig:confusionHiNERCollapsed}. We observe that majority of the errors involve tagging a named entity as not a named entity followed by boundary errors \textit{i.e.,} mislabelling \textit{B-Person} with \textit{I-Person} and vice-versa. Also \textit{Organization} entities tends to be confused with \textit{Location} entities. The majority of the errors are produced when the model cannot identify a token as a name itself. Additionally, we report a detailed analysis of the types of errors made by the XLM-R\textsubscript{large} Model on HiNER data. Table \ref{tab:analysisDetailedHiNER} provides more detailed insights into the performance of the system by reporting \textit{strict}, \textit{exact} evaluation metrics \cite{chinchor-sundheim-1993-muc}. We use the \textit{nervaluate} package\footnote{\url{https://github.com/MantisAI/nervaluate}} to calculate the above statistics for each entity type. Specifically, we pick the predictions from one of the runs using \textit{XLM-R\textsubscript{large}} as this model consistently gave better results compared to the other pre-trained language models. We use two different evaluation schemas mentioned in the Table \ref{tab:evaluationSchema}. \textit{Exact} encourages models to identify the named entity phrase correctly while ignoring the type mismatch.

We observe that for some entity type like \textit{Location, NUMEX, Organization} \textit{Missed} errors are more than the \textit{Spurious} errors. On the other hand, for entity types like \textit{Person, Misc, Language, Game, TIMEX} \textit{Spurious} errors are more. We additionally report F1-Score according to the evaluation schema for each entity type. The most challenging entity categories are \textit{Literature}, \textit{Festival}, \textit{MISC}, \textit{Language}, \textit{Religion}, and \textit{Game} entities. We observe that the model is able to identify \textit{Misc, Language, Religion, Literature} as named entities but unable to assign the correct entity type. This cab seen in the F-Score different between \textit{Strict} and \textit{Exact} evaluation schema. \newline

\begin{table}[!hb]
\begin{tabular}{@{}lp{5cm}@{}}
\toprule
\multicolumn{1}{c}{\textbf{\begin{tabular}[c]{@{}c@{}}Evaluation\\ Schema\end{tabular}}} & \multicolumn{1}{c}{\textbf{Explanation}} \\ \midrule
Strict & The exact boundary surface string match and entity type match \\
Exact & The exact boundary match over the surface string, regardless of the type \\
\bottomrule
\end{tabular}
\caption{Short Description of the Evaluation Schema used}
\label{tab:evaluationSchema}
\end{table}

For each type of evaluation schema (\textit{i.e., strict and exact}) we report the following categories of errors listed in Table \ref{tab:errorCategories}.

\begin{table}[!hb]
\begin{tabular}{@{}lp{5cm}@{}}
\toprule
\multicolumn{1}{c}{\textbf{Error type}} & \multicolumn{1}{c}{\textbf{Explanation}} \\ \midrule
Correct & The gold annotations and the system predictions are the same \\
Incorrect & The system prediction and the gold annotation don’t match \\
Missing & The system prediction classifies an entity as not a named entity \\
Spurius & The system prediction classifies a non named-entity as a named entity \\ \bottomrule
\end{tabular}
\caption{Short Description of the Categories of Errors}
\label{tab:errorCategories}
\end{table}

\begingroup\small
\bottomcaption{Detailed Strict and Exact Results on HiNER data from XLM-R\textsubscript{large} Model} 
\label{tab:analysisDetailedHiNER}
\centering
\tablefirsthead{%
\toprule
 & \textbf{Error Category} & \textbf{Strict} & \textbf{Exact}\\ 
\midrule}
\tablehead{%
\multicolumn{2}{@{}l}{Table \thetable, continued}\\
\midrule[\heavyrulewidth]
 & & \textbf{Strict} & \textbf{Exact}\\ 
\midrule}
\tabletail{\bottomrule}
\tablelasttail{\bottomrule}
    \begin{xtabular}{llrr}
    \multirow{5}{*}{Person} & Correct &    6475 &   6565 \\
         & Incorrect &     622 &    532 \\
         & Missed &     427 &    427 \\
         & Spurious &     537 &    537 \\
         \cmidrule{2-4}
         & F1 &  0.8543 & 0.8662 \\
         \midrule
\multirow{5}{*}{Location} & Correct &   37960 &  38113 \\
         & Incorrect &    1028 &    875 \\
         & Missed &    1199 &   1199 \\
         & Spurious &     688 &    688 \\
         \cmidrule{2-4}
         & F1 &  0.9506 & 0.9545 \\
         \midrule
\multirow{5}{*}{NUMEX} & Correct &    3047 &   3097 \\
         & Incorrect &     567 &    517 \\
         & Missed &    1048 &   1048 \\
         & Spurious &     526 &    526 \\
         \cmidrule{2-4}
         & F1 &  0.6923 & 0.7037 \\
         \midrule
\multirow{5}{*}{Organization} & Correct &    4195 &   4263 \\
         & Incorrect &     535 &    467 \\
         & Missed &     626 &    626 \\
         & Spurious &     544 &    544 \\
         \cmidrule{2-4}
         & F1 &  0.7893 & 0.8021 \\
         \midrule
\multirow{5}{*}{Misc} & Correct &     804 &    882 \\
         & Incorrect &     137 &     59 \\
         & Missed &     139 &    139 \\
         & Spurious &     265 &    265 \\
         \cmidrule{2-4}
         & F1 &  0.7034 & 0.7717 \\
         \midrule
\multirow{5}{*}{Language} & Correct &    1115 &   1133 \\
         & Incorrect &      60 &     42 \\
         & Missed &      15 &     15 \\
         & Spurious &      42 &     42 \\
         \cmidrule{2-4}
         & F1 &  0.9265 & 0.9414 \\
         \midrule
\multirow{5}{*}{Game} & Correct &     276 &    279 \\
         & Incorrect &      57 &     54 \\
         & Missed &      36 &     36 \\
         & Spurious &     145 &    145 \\
         \cmidrule{2-4}
         & F1 &  0.6517 & 0.6588 \\
         \midrule
\multirow{5}{*}{TIMEX} & Correct &    3018 &   3055 \\
         & Incorrect &     328 &    291 \\
         & Missed &     307 &    307 \\
         & Spurious &     363 &    363 \\
         \cmidrule{2-4}
         & F1 &  0.8199 & 0.8299 \\
         \midrule
\multirow{5}{*}{Religion} & Correct &     175 &    183 \\
         & Incorrect &      26 &     18 \\
         & Missed &      33 &     33 \\
         & Spurious &      32 &     32 \\
         \cmidrule{2-4}
         & F1 &  0.7495 & 0.7837 \\
         \midrule \\
         \\
\multirow{5}{*}{Literature} & Correct &     104 &    116 \\
         & Incorrect &      35 &     23 \\
         & Missed &      42 &     42 \\
         & Spurious &      21 &     21 \\
         \cmidrule{2-4}
         & F1 &  0.6100 & 0.6804 \\
         \midrule
\multirow{5}{*}{Festival} & Correct &      26 &     26 \\
         & Incorrect &       6 &      6 \\
         & Missed &       8 &      8 \\
         & Spurious &      15 &     15 \\
         \cmidrule{2-4}
         & F1 &  0.5977 & 0.5977 \\
    \end{xtabular}
\endgroup

\vspace{0.3cm}

\section{Conclusion and Future Work}

We describe our efforts to create a sizeable human-annotated dataset, HiNER, for the task of Named Entity Recognition in the Hindi language. We discuss the motivation for this research, the challenges specific to Hindi NER, and provide coverage of the past research performed for the NER task in Hindi. We discuss the dataset creation in detail and provide an in-depth analysis of the tag-set used to label our NER data. We also describe the NER annotation tool created to help our annotators along with the NER engine it utilises to label the data initially on the tool interface. We split our data and performed experiments to evaluate different language models to perform the NER task by fine-tuning them. We also perform similar experiments on another dataset for a comparative evaluation. We discuss our results in detail and show how large human-annotated NER data is essential for the task of Hindi NER. We release this dataset and the models we train; for the NLP community to utilise them for the downstream NLP tasks. We choose the CC-BY-SA 4.0 Licensing terms to release this data. In future, we plan to keep extending this dataset with the help of our ongoing annotation process.


\section*{Bibliographical References}\label{reference}

\bibliographystyle{lrec2022-bib}
\bibliography{lrec2022-example}



\end{document}